\documentclass[conference]{IEEEtran}
\IEEEoverridecommandlockouts
\usepackage{cite}
\usepackage{amsmath,amssymb,amsfonts}
\usepackage{algorithmic}
\usepackage{graphicx}
\usepackage{textcomp}
\usepackage{xcolor}
\usepackage{subcaption}
\usepackage{ulem}
\usepackage{comment}

\usepackage{float} 

\def\BibTeX{{\rm B\kern-.05em{\sc i\kern-.025em b}\kern-.08em
    T\kern-.1667em\lower.7ex\hbox{E}\kern-.125emX}}
\begin{document}

\newcommand{\ar}[1]{\textcolor{blue}{#1}}
\newcommand{\as}[1]{\textcolor{olive}{ #1}}
\newcommand{\arr}[1]{\textcolor{red}{\textit{Razi: #1}}}

\title{Enhanced Cooperative Perception for Autonomous Vehicles Using Imperfect Communication\\
\thanks{This material is based upon work supported by the National Science Foundation under Grant Number CNS2204721 and MIT Lincoln Lab under Grant Number 7000565788.}
}

\author{\IEEEauthorblockN{1\textsuperscript{st} Ahmad Sarlak}
\IEEEauthorblockA{\textit{School of computing)} \\
 \textit{Clemson University}\\
 Clemson, SC, USA\\
 asarlak@clemson.edu}
 \and
 \IEEEauthorblockN{2\textsuperscript{nd} Hazim Alzorgan}
\IEEEauthorblockA{\textit{School of computing)} \\
 \textit{Clemson University}\\
 Clemson, SC, USA\\
 alzorg@clemson.edu}
 \and
 \IEEEauthorblockN{3\textsuperscript{rd}  Sayed Pedram Haeri Boroujeni}
\IEEEauthorblockA{\textit{School of computing)} \\
 \textit{Clemson University}\\
 Clemson, SC, USA\\
 shaerib@clemson.edu}
 \and
 \IEEEauthorblockN{4\textsuperscript{th}  Abolfazl Razi}
\IEEEauthorblockA{\textit{School of computing)} \\
 \textit{Clemson University}\\
 Clemson, SC, USA\\
 arazi@clemson.edu}
 \and
 \IEEEauthorblockN{5\textsuperscript{th} Rahul Amin}
\IEEEauthorblockA{\textit{Lincoln Laboratory)} \\
 \textit{Massachusetts Institute of Technology}\\
 Lexington, MA, USA\\
 rahul.amin@ll.mit.edu}
}

\author{
    \IEEEauthorblockN{
        Ahmad Sarlak\IEEEauthorrefmark{1},
        Hazim Alzorgan\IEEEauthorrefmark{1},
        Sayed Pedram Haeri Boroujeni\IEEEauthorrefmark{1},
        Abolfazl Razi\IEEEauthorrefmark{1},
        Rahul Amin\IEEEauthorrefmark{2}
    }
    \IEEEauthorblockA{
        \IEEEauthorrefmark{1}School of Computing, Clemson University, Clemson, SC, USA \\
        Email: \{asarlak, halzorg, shaerib, arazi\}@clemson.edu
    }
    \IEEEauthorblockA{
        \IEEEauthorrefmark{2}Lincoln Laboratory, Massachusetts Institute of Technology, Lexington, MA, USA \\
        Email: rahul.amin@ll.mit.edu
    }
}

\vspace{4 cm}

\maketitle

\begin{abstract}

Sharing and joint processing of camera feeds and sensor measurements, known as Cooperative Perception (CP),  has emerged as a new technique to achieve higher perception qualities. CP can enhance the safety of Autonomous Vehicles (AVs) where their individual visual perception quality is compromised by adverse weather conditions (haze as foggy weather), low illumination, winding roads, and crowded traffic. While previous CP methods have shown success in elevating perception quality, they often assume perfect communication conditions and unlimited transmission resources to share camera feeds, which may not hold in real-world scenarios. Also, they make no effort to select better helpers when multiple options are available.

To cover the limitations of former methods, in this paper, we propose a novel approach to realize an optimized CP under constrained communications.   
At the core of our approach is recruiting the best helper from the available list of front vehicles to augment the visual range and enhance the Object Detection (OD) accuracy of the ego vehicle. In this two-step process, we first select the helper vehicles that contribute the most to CP based on their visual range and lowest motion blur. Next, we implement a radio block optimization among the candidate  
vehicles to further improve communication efficiency. We specifically focus on pedestrian detection as an exemplary scenario. To validate our approach, we used the CARLA simulator to create a dataset of annotated videos for different driving scenarios where pedestrian detection is challenging for an AV with compromised vision. Our results demonstrate the efficacy of our two-step optimization process in improving the overall performance of cooperative perception in challenging scenarios, substantially improving driving safety under adverse conditions. Finally, we note that the networking assumptions are adopted from LTE Release 14 Mode 4 side-link communication, commonly used for Vehicle-to-Vehicle (V2V) communication. Nonetheless, our method is flexible and applicable to arbitrary V2V communications

\end{abstract}


\section{Introduction}

Cooperative Perception (CP) represents a new paradigm in Autonomous Vehicles (AVs) by mitigating the limitations of individual vehicle perception \cite{chen2022milestones, boroujeni2024comprehensive}. CP involves sharing and the integrative processing of camera feeds and sensor readings by multiple AVs within a network, enabling them to collectively enhance their situational awareness. 
The use of CP stems from the fact that the reliable operation of AVs can face significant challenges due to perception errors in adverse weather conditions and complex traffic scenarios, where individual vehicle perception systems may be compromised. CP has recently emerged as a promising solution to enhance AV safety by leveraging collaborative information sharing among vehicles.

\begin{figure}[H]
\begin{subfigure}{\columnwidth}
  \includegraphics[width=\textwidth]{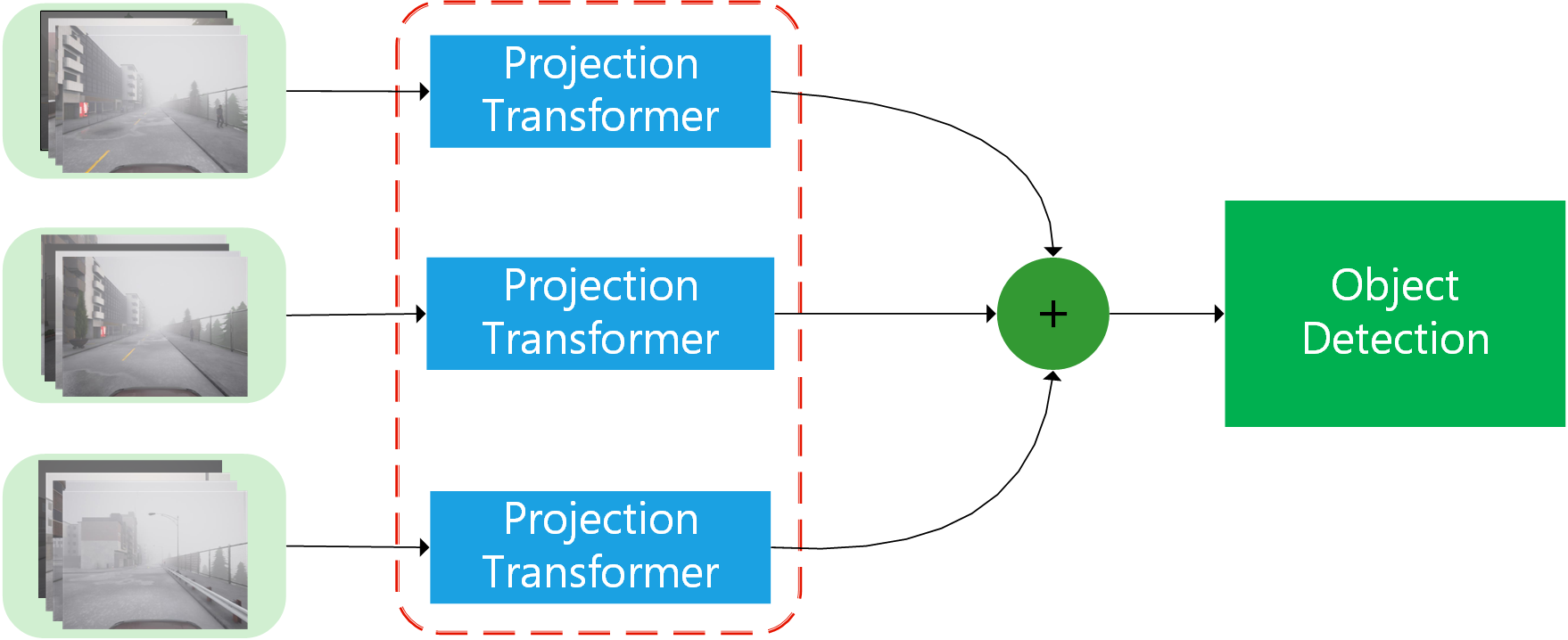}
  \caption{}
\end{subfigure}

\begin{subfigure}{\columnwidth}
  \includegraphics[width=\textwidth]{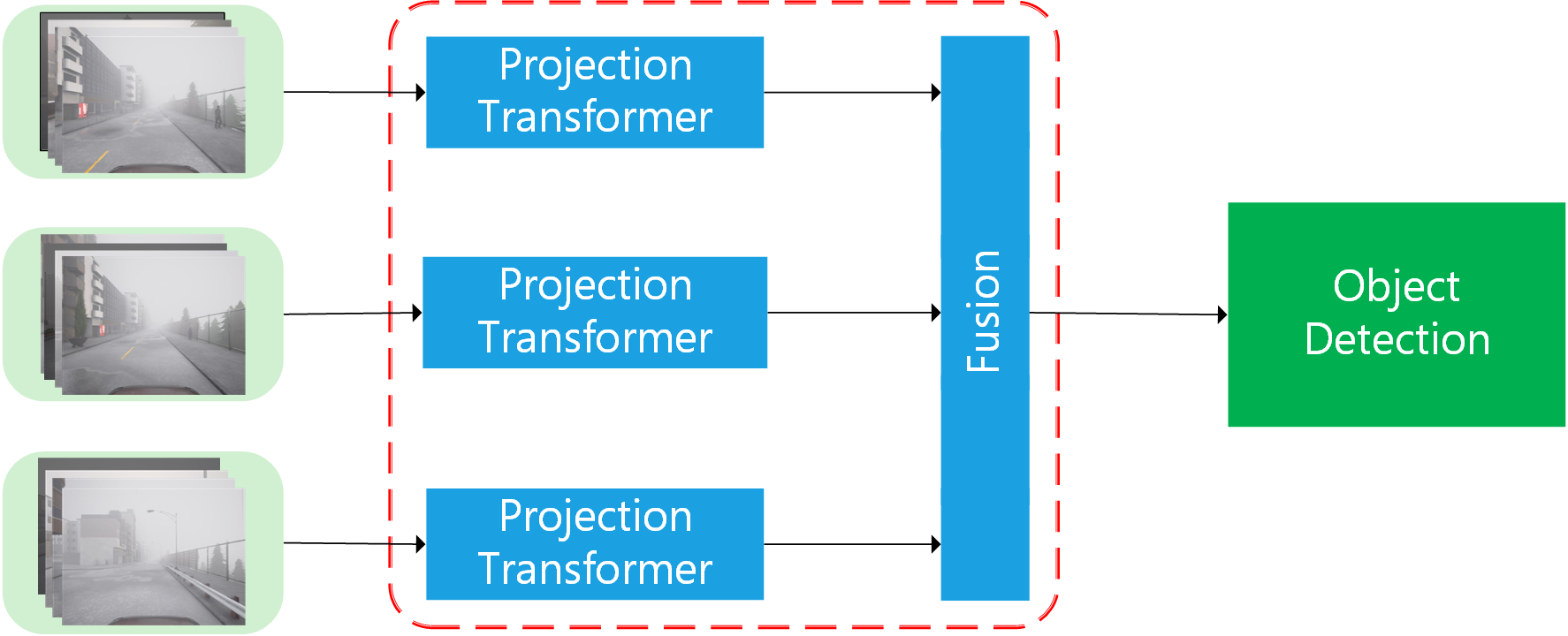}
  \caption{}
\end{subfigure}

\begin{subfigure}{\columnwidth}
  \includegraphics[width=\textwidth]{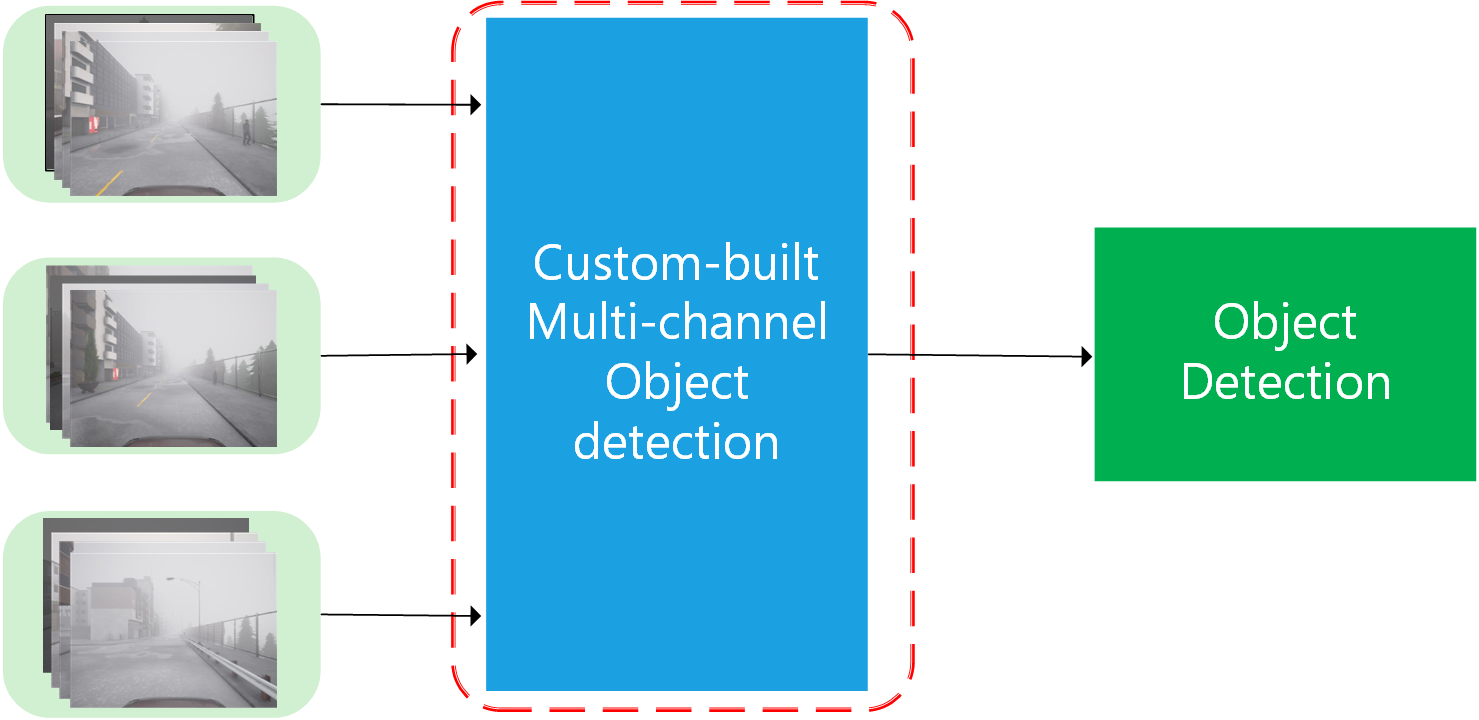}
  \caption{}
  \end{subfigure}
\vspace{-0.5cm}
\caption{Three fusion approaches used in cooperative perception, including (a) early fusion, (b) late fusion, and (c) integrative analysis.}
\label{fig:fusion}
\vspace{-0.2 in}
\end{figure}

A typical CP approach employs Vehicle-to-Vehicle(V2V) communication technology to share sensory data among vehicles for collective processing \cite{xu2022opv2v, xu2022cobevt, sarlak2023diversity}. To realize the collective processing of shared visual information, there exist three mainstream fusion methods, as illustrated in Figure \ref{fig:fusion}. The first approach is \textit{early fusion}, where the raw captured images by multiple vehicles are translated to the same field of view through projection transformation, and then mixed to obtain a single high-quality image to be used by the downstream learning-based tasks such as object detection (OD), depth estimation, reinforcement learning (RL), etc. (Fig. \ref{fig:fusion} (a)). The main issue of this approach is the sensitivity of the fused image to the parameters of projection transformation. 
The second approach is \textit{late fusion}, where each image is first processed by the learning-based task, then the obtained results are mixed (Fig. \ref{fig:fusion} (b)). For instance, one may execute OD on individual images to obtain object classes and bounding boxes, to be fused to get more precise results. The advantage of these two methods is their easy integration with existing deep learning (DL)-based visual processing methods \cite{boroujeni2024ic}. The third approach is more intricate and involves developing new DL \textit{architectures} that jointly process multiple image/video channels(Fig. \ref{fig:fusion} (c)). This method can yield better results for specific tasks but requires custom-designed DL architectures that can limit their applicability. In this work, we use the \textit{late fusion} method for its simplicity and practicality.

While existing CP methods have demonstrated success in addressing these challenges, most of them have taken the unrealistic assumptions of perfect communication conditions and unlimited transmission resources \cite{xu2022opv2v, xu2022cobevt}, which may not align with the real-world scenarios. To the best of our knowledge, there exist only two works that consider imperfect communications in CP. First is \cite{xu2022v2x}, where the authors present a vision transformer, V2X-ViT, along with a CP framework that utilizes Vehicle-to-Everything (V2X) communication to improve AV perception in terms of 3D object detection. While the authors try to address V2X challenges, their primary focus is on communication delays, overlooking other essential communication aspects (packet drop rate, throughput, ...). Their proposed architecture aimed to enhance robustness to delay in the network.

Likewise, \cite{li2023learning} intends to address the impact of imperfect V2X  communication in cooperative perception for 3D object detection. They proposed an innovative approach with an LC-aware Repair Network (LCRN) and a V2V Attention Module (V2VAM) to effectively mitigate the adverse effects of imperfect communication. They significantly improved the detection performance demonstrated on the OPV2V dataset. However, they only tried to adjust their method for varying channel quality without incorporating channel quality into the vehicle selection process.

We propose a novel approach to CP to overcome the limitations of the previous methods by enabling selective communication, where the helper vehicles are selected based on their visual information as well as their channel quality. The networking parameters are adopted from the LTE Release 14 Mode 4 side-link communication used for V2V communication. 
By doing so, we achieve enhanced situational awareness through higher CV-based detection accuracy under imperfect and constrained communication scenarios, compared to CV methods that blindly or randomly selected helper vehicles. 
In this work, we select helper vehicles based on their contribution to extending the visual range of the ego vehicle, considering their geo-locations as well as their imperfect channel conditions. Additionally, we take into account the vehicles' relative velocities, because the motion blur impacts their imaging quality \cite{xiao2021vehicle, cortes2018velocity, ye2020federated} and consequently the ultimate accuracy of the developed CP-based object detection accuracy.

\begin{figure}[H]
\begin{subfigure}{0.7\columnwidth}
  \includegraphics[width=\textwidth]{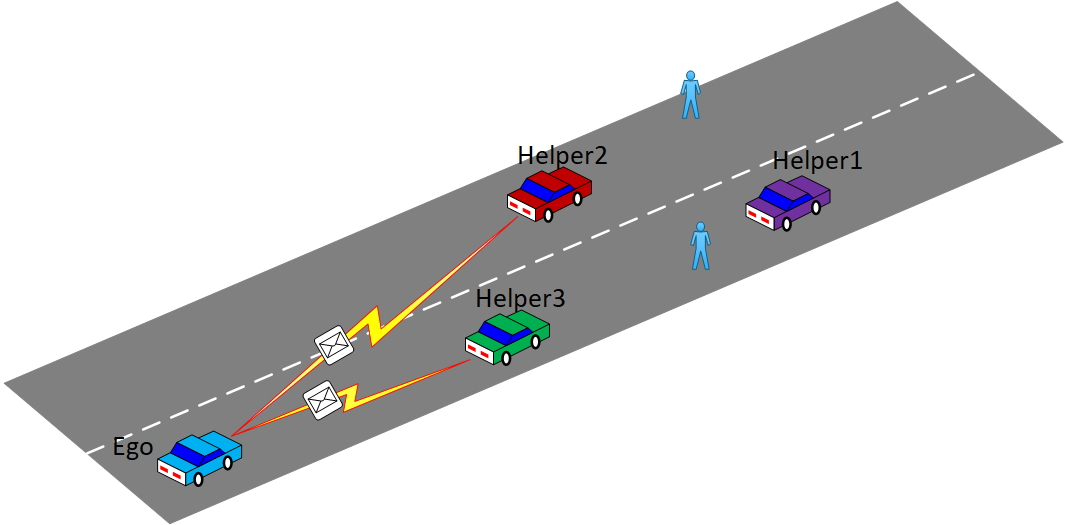}
  \caption{}
\end{subfigure}

\begin{subfigure}{0.7\columnwidth}
  \includegraphics[width=\textwidth]{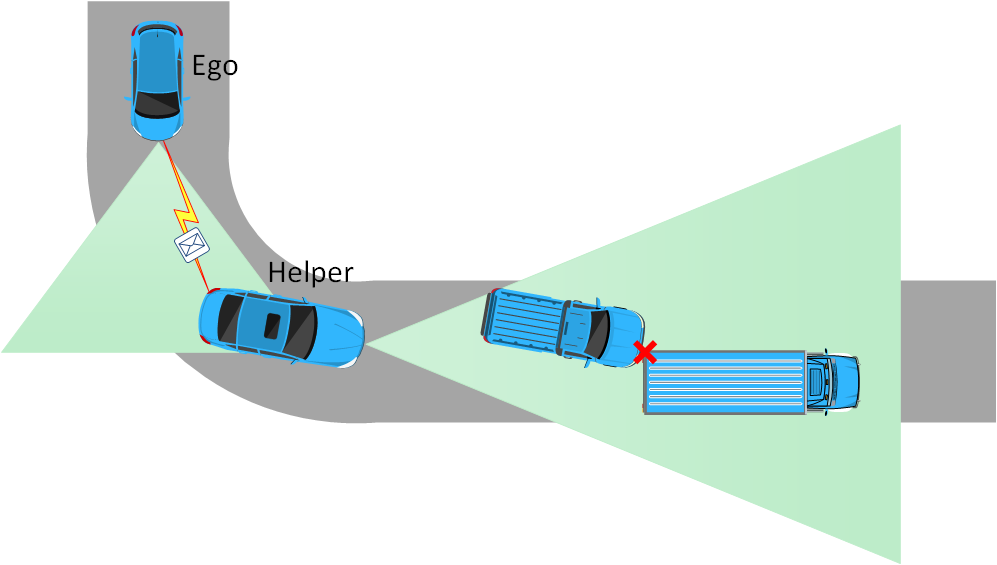}
  \caption{}
\end{subfigure}

\caption{Two scenarios that may occur on the road are: (a) in foggy weather, helper vehicles closer to pedestrians (Green and Purple) can share their cameras with the ego vehicle (Red vehicle), and (b) on a curved road, a helper can detect an accident and share its camera with the ego vehicle.+}
\label{fig:scenarios}
\vspace{-0.2 in}
\end{figure} 
To adopt more realistic assumptions, we note that transmission errors in C-V2X mode 4 arise from different factors, including (1) using half-duplex transmission, (2) receiving signal strengths below the sensing threshold, (3) propagation effects, and (4) packet collisions \cite{gonzalez2018analytical}. 
In addition to considering channel quality in selecting the best helper vehicles, we also enhance the resulting communication efficiency through radio block optimization among the selected vehicles\cite{chinipardaz2024joint}.

To investigate the effectiveness of the proposed approach, we concentrate on pedestrian detection as an exemplary scenario. However, the proposed framework is generic and can be used for other applications, such as speed estimation, lane detection, vehicle classification, traffic sign detection, traffic light interpretation, etc. Through the generation of frames using the CARLA simulator \cite{dosovitskiy2017carla}, annotation of data, and compilation of a comprehensive dataset, we conduct experiments to validate the performance of our proposed method.

The results obtained from our two-step optimization process, present notable improvements in CP by selecting the best helper to increase the ego vehicle's visual range. 

The rest of the paper is organized as follows. In section 2, a more inclusive review of related work is presented. In section 3, the system model is presented. Finally, the proposed method is evaluated through intensive simulations in section 4, followed by concluding remarks in section 5.

\section{Related Works}
\subsection{CAVs with Imperfect Communications}
V2V LTE Release 14 presents advancements in V2V communication, enhancing connectivity and communication protocols for improved safety and efficiency in vehicular networks. However, there are limitations in V2V communications, that that should be considered when developing networked service provisioning systems. Authors in \cite{gonzalez2018analytical} evaluate the communication performance of C-V2X or LTE-V Mode 4, focusing on average Packet Delivery Ratio (PDR)
and four types of transmission errors, with validating various transmission parameters and traffic densities. Authors in \cite{chen2020joint, li2020federated} highlight the importance of wireless factors, such as packet errors and limited bandwidth in Federated Learning (FL) implementations. Li et al. \cite{li2022federated} presents the importance of efficient resource allocation techniques in V2V communication networks in FL, particularly in the vehicular safety services area, where delay and reliability requirements need optimal utilization of limited spectrum resources. The authors state that such resource allocation strategies are essential to mitigate interference, support various quality-of-service requirements, and ensure the success of emerging vehicle-related services in dynamic and fast-changing vehicular environments.
These researches show that imperfect communication can play a key role in CP; therefore, it should not be overlooked when forming a network of vehicles for collective perception. 

\subsection{3D Object Detection}

Our target application, in this work, is 3D OD, which
involves identifying and locating objects in three-dimensional space.  
It is an integral part of autonomous driving and crash avoidance \cite{li2023learning}.
3D OD can be applied to different perception domains, including visions-based (e.g., regular and IR cameras), sensor-based (e.g., RADAR/LiDAR), and hybrid methods, with the following notable implementations. 

Reading et al. \cite{reading2021categorical} proposed Categorical Depth Distribution Network (CaDDN), as a monocular vision-based 3D OD method. This method utilizes predicted categorical depth distributions for each pixel to enhance depth estimation accuracy and achieve a top-ranking performance on the KITTI dataset, showcasing its effectiveness in addressing the challenges of monocular 3D detection. Authors of \cite{wang2022detr3d} introduced a multi-camera 3D OD framework that operates directly in 3D space, leveraging sparse 3D object queries to index 2D features from multiple camera images.
 \cite{shi2019pointrcnn} 
proposes PointRCNN which is a 3D OD framework, where using point cloud data directly to generate  high-quality 3D proposals instead of generating proposals from RGB image
or projecting point cloud to bird’s view or voxels.

An exemplary implementation of LiDAR-based OD methods is PointPillars, a highly effective point cloud encoder, that employs PointNets to organize point clouds into vertical columns (pillars), demonstrating superior speed and accuracy \cite{lang2019pointpillars}. 
They outperform existing methods on the KITTI benchmarks while achieving a higher accuracy in OD in comparison to \cite{yan2018second, xiang2017subcategory, ku2018joint}. Finally, a hybrid camera-LiDAR fusion detection method, called RCBEV, is proposed in \cite{zhou2023bridging}. RCBEV implements a radar-camera feature fusion method for 3D OD in autonomous driving using nuScenes and VOD dataset through an efficient top-down feature representation and a two-stage fusion model which bridges
the view disparity between radar and camera features.
\vspace{-0.1cm}

\subsection{Fusion Methods}
The information of multiple perception systems can be collectively processed in three different modes, namely \textit{early}, \textit{integrative}, and \textit{late fusion}. The authors of \cite{chen2019cooper} introduce an \textit{early fusion} approach enabling cooperative perception through the fusion of LiDAR 3D point clouds from diverse positions, improving detection accuracy and expanding the effective sensing area. They validated their method using KITTI and T\&J dataset, demonstrating the feasibility of transmitting point cloud data via existing vehicular network technologies. 
Qiu et al. present an infrastructure-less CP system, leveraging direct vehicle-to-vehicle communication to efficiently share sensor readings while optimizing transmission schedules to enhance safety in dense traffic scenarios \cite{qiu2021autocast}.

A \textit{late fusion} machine approach is adopted in \cite{rawashdeh2018collaborative} to enhance the accuracy of shared information for collaborative automated driving. First, they employed Convolutional Neural Networks (CNN) for OD and classification to extract positional and dimensional information. Next, they combined extracted information to 
enhance driving safety 
in V2V collaborative systems. Article \cite{arnold2020cooperative} explores CP for 3D OD using two fusion schemes, \textit{early} and \textit{late fusion}, showing that 
\textit{early fusion} outperforms \textit{late fusion} in challenging scenarios, 
while authors in \cite{yu2022dair} introduce the DAIR-V2X dataset for Vehicle-Infrastructure Cooperative Autonomous Driving (VICAD) and formulates the Vehicle-Infrastructure Cooperative 3D OD (VIC3D) problem, proposing the Time Compensation \textit{late fusion} framework as a benchmark, highlighting the importance of collaborative solutions and addressing challenges in autonomous driving.

Both \textit{early} and \textit{late fusion} methods try to combine information either before or after applying the target DL-based application, like 3D OD. This brings the convenience of deploying existing methods. Recently, some attempts have been made to develop integrative processing of multi-channel inputs to achieve elevated performance. 
Xu et al. \cite{xu2022v2x} introduced a fusion model V2X-ViT, based on the popular vision transformer architecture equipped with heterogeneous multi-agent and multi-scale window self-attention modules for robust CP in autonomous vehicles, 
achieving state-of-the-art 3D OD performance in challenging and noisy environments using a large-scale V2X perception dataset. 
Another \textit{integrative} end-to-end DL-based architecture is presented in \cite{cui2022coopernaut} 
that utilizes cross-vehicle perception with LiDAR data encoded into compact point-based representations, demonstrating an improvement in average success rate over egocentric driving models in challenging scenarios, with a 5× smaller bandwidth requirement compared to prior work, as evaluated in the AUTOCASTSIM simulation framework.


\section{System Model}
In our approach, the ego vehicle sends request-for-help messages to front vehicles when individual perception quality is not satisfactory. We assume that  
$N$ vehicles, $V_1$, $V_2$, \dots, $V_N$, notify their intention to provide CP assistance through Ack messages along with their position ($x_i$, $y_i$), speed $v_i$, channel conditions, and sample images.  Additional details, such as the distance to the ego vehicle, approximate vision range, motion blur, field of view, required transmission energy budget $E$, and communication errors in the C-V2X mode 4, are estimated based on the gathered information (position, channel conditions) and refined through an examination of the sample images. Then, various error factors, including half-duplex transmission, received signal power below the sensing power threshold, propagation effects, and packet collisions, are combined into $\beta$, denoting the effective packet error probability.
Once the information is obtained, the ego vehicle selects $M$ out of $N$ volunteered vehicles that collectively achieve the highest performance under given conditions.

The main objective of selecting helpers is to maximize contextual information diversity through an effective and informed choice of helpers. In our case, the key contributing factors include the vehicle's contribution to extending the visual range of the ego vehicle (based on their positions), the quality of their image based on motion blur (based on their velocities), and the channel conditions. We develop a set of objective functions, as elaborated below, that quantify these factors and facilitate selecting the best helper. 
\vspace{-0.1cm}

\begin{figure}[htbp]
\begin{center}
\centerline{
\includegraphics[width=1\columnwidth]{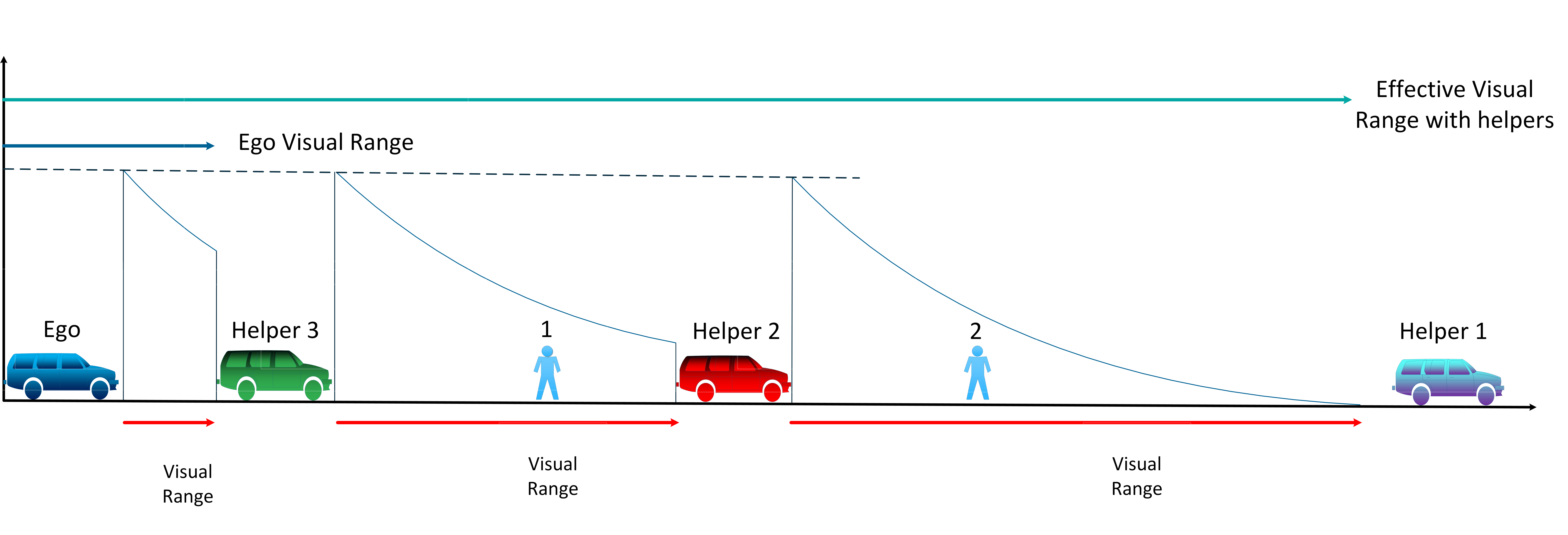}}
\vspace{-0.3cm}
\caption{The individual visual range for vehicles. Recruiting helpers extend the visual range of the ego vehicle through cooperative perception.}
\label{fig:vs1}
\end{center}
\end{figure} 
\vspace{-0.1cm}

\subsection{Approximate Vision Range}
A key factor in selecting helpers is the collective visual range of the ego vehicle that enables accurate object detection. The visual range of each vehicle varies with weather conditions, light conditions, and the presence of obstacles. The effective individual visual range of vehicles is demonstrated in Fig. \ref{fig:vs1}.

By optimizing the selection process, we ensure that the chosen helpers effectively contribute to expanding the collective visual range of the ego vehicle.

We assume that all vehicles know their location $(x_i, y_i)$ and their distance to the front obstacle, and share this information with the ego vehicle. When calculating the total visual range, we use function $h(x)$ to determine individual visual ranges
\vspace{-0.1cm}
\begin{align}
    \nonumber
    &h(x) = u [ x- (x_{i}+ L_{i})]-u[x-x_{i}],\\
    &L_i = \min (T, x_{i+1} - x_{i}),
\end{align}
\vspace{-0.1cm}
where $x_i$ represents the Longitudinal motion along the linear road on x-axis. Otherwise, we substitute $x\leftarrow x \ cos \theta $ or $x\leftarrow y \ sin \theta $ from positions if the road makes $\theta$ with x-axis. Here, $h_i(x)$ is a rectangular pulse of length $L_i$, representing the $i^{th}$ vehicle's visual range until the threshold $T$ defined by the weather and light conditions or the distance to the front vehicle $x_{i+1}-x_i$, whichever is shorter. The sum of these pulses from selected vehicles represents the augmented visual range.  

In addition to expanding the visual range, we desire to prioritize selecting closer vehicles. This is crucial because having a clear vision for detecting closer objects is imperative for safe driving. 
For instance, in Fig. \ref{fig:vs1} Helper 3 can help with detecting both pedestrians at least partially, Helper 2 can help only with detecting one pedestrian, and Helper 1's contribution is irrelevant (not useful) in detecting either of pedestrians.

To capture this effect, we use an exponential decay function $g(x)= e^{-a x}$ with a tuning parameter $a$ to weigh the augmented visual range to promote selecting closer helpers. If the binary vector $\vec{\alpha}=(\alpha_1,\alpha_2,\cdots, \alpha_N)$  represents the volunteer vehicles ($\alpha_i=1$: selected, $\alpha=0$: not selected), the weighted augmented visual range of selected vehicles can be expressed as
\vspace{-0.3cm}
\begin{align}
    f_1(\vec{\alpha}) = 
    \int_{x=0}^{\infty} g(x) \cdot\left[\sum_{i=1}^N \alpha_i h_i(x)\right].
\end{align}

\subsection{Motion Blur Level}
We assume that all vehicles are equipped with 
similar cameras with an equal resolution and field of view. In this paper, the only important parameter affecting motion blur is the speed of the vehicles. 
Motion blur represents the concept of the decline in image quality for higher speeds. We follow the characterization of motion blur in \cite{xiao2021vehicle} to obtain
\vspace{-0.1cm}
\begin{align}
f_2(\vec{\alpha})=
\sum_{i=1}^N \alpha_i 
 \frac{v_i^{\prime} T[f \cos (\phi)-Q S \sin (\phi)]}{v_i^{\prime} T Q \sin (\varphi)+z p},
\end{align}
for the vehicles joining CP ($\alpha_i=1$). Here, $T$ is the exposure time interval, $f$ is the camera focal length, $Q$ is the charge-coupled device (CCD) pixel size in the horizontal direction, $S$ is the starting position of the object in the image (in pixels), $\phi$ is the angle between the motion direction and the image plane, $d$ is the perpendicular distance from the starting point of the moving object to the pinhole \cite{cortes2018velocity}.
The effect is maximized and becomes linear in velocity, when the image plane is parallel to the motion direction ($\phi = 0$), as we get
\begin{align} 
f_2(\vec{\alpha})=
\sum_{i=1}^N \alpha_i 
 \frac{v_i^{\prime} T f }{z p}.
\end{align}

\subsection{Communications Parameters}
 
The transmission energy budget and packet drop rate are essential metrics in C-V2X communications, as they directly impact the reliability and effectiveness of V2X communication and the overall safety and efficiency of AVs by determining the successful exchange of critical information among vehicles and infrastructure. 

For the definition of effective throughput, we absorb the impact of all error terms in LTE C-V2X into packet error rate $\beta$. As a result, the effective throughput $\zeta_i$ of vehicle $V_i$ is 
\begin{align}
\zeta_i= R_{ch}/E[R_i] = R_{ch} (1 - \beta_i) w_i,
\end{align}
where $R_{ch}$ is the rate of channel, and $w_i$ is the bandwidth of the resource block assigned to vehicle $i$. 
Here, $R_i$ is the number of re-transmissions with the following expected value
\begin{align}
E[R_i]=1/(1 - \beta_i).
\end{align}

Therefore, the average throughput under selection vector $\alpha$ is 
\begin{equation} \label{eq:th}
f_3(\boldsymbol{w})= \sum_{i=1}^{\mathrm{N}} w_i\zeta_i=  \sum_{i=1}^{\mathrm{N}} R_{ch} (1 - \beta_i) w_i .
\end{equation}

\begin{equation} \label{eq:th}
f_3(\vec{\alpha}, \vec{w})= \sum_{i=1}^{\mathrm{N}} \alpha_i\zeta_i=  \sum_{i=1}^{\mathrm{N}} R_{ch} (1 - \beta_i) w_i \alpha_i.
\end{equation}

The transmission energy of vehicle $V_i$ is
\begin{align}
    e_i = \frac{P_i^{t r}}{R_i \cdot\left(1-\beta_i\right)}\alpha_i
\end{align}
where $p_i^{t r}$ denotes the average transmission power, considering the number of re-transmissions represented by $1/(1-\beta_i)$. Therefore, the average energy consumption for selected vehicles is 
\vspace{-0.1cm}

\begin{align}
f_4(\vec{\alpha}, \vec{w})= \sum_{i=1}^{\mathrm{N}} \alpha_i   \frac{P_i^{t r}}{R_{c h}\left(1-\beta_i\right)},
\end{align}

One may also include then End-to-End (E2E) delay in the optimization. In general, E2E delay accounts for sampling and perception delays in the sender, queuing delays, channel setup delays, congestion
and re-transmissions, actual transmission delays, as well as processing delays in the receiver \cite{cruz1991calculus}, We take a simplistic assumption and model $D_i^{(1)}$, the E2E delay for one packet for vehicle $V_i$ as an exponentially distributed continuous-valued Random Variable (RV) with vehicle-specific mean $\lambda_i$ following some prior work \cite{razi2017delay}, and then specifically, we define

\begin{align}   \label{eq:delay}
\nonumber
&D_i^{(1)}\sim f_{\lambda_i}(d_i), \\
&f_{\lambda_i}(d_i)= 
\begin{cases}
\lambda_i e^{-\lambda_i d_i}, & d_i\geq 0, \\ 0, & d_i<0.
\end{cases}
\end{align}

Here, we use capital letters for RVs and lowercase letters for their realizations. We consider that an average delay $E[D_i^{(1)}]=1 / \lambda_i$ for each vehicle $V_i$ remains constant in one transmission slot. 
Here, we consider $D_i^{(1)}$ captures all delay terms. However, if it accounts only for queuing delay, we can add the constant term $D_{tr}=l/ R_{ch}$ accounting for the actual transmission delay, where $l$ is the packet length (in bits). For convenience, we exclude delay in the optimization, since CP requires near-realtime synchronization between packets from helper vehicles.
\vspace{-0.1cm}
\subsection{Optimization Problem}
\vspace{-0.1cm}
With both the visual range and the speed of each vehicle, a selection problem arises. This process aims to select the vehicle that not only extends the overall visual coverage for the ego vehicle but also contributes to capturing higher-quality images with reduced motion blur. 

 We formulated our problem as two optimizations, in which first, we tried to select the best $M$ vehicles out of $N$ and then allocate them among the selected vehicles. Our optimization is based on a set of transmission Key Performance Indicators (KPIs) calculated for selected vehicles. Specifically if the binary vector $\vec{\alpha}= (\alpha_1, \alpha_2, \cdots, \alpha_N)$ represents the selected vehicles.

The optimization problem is
\vspace{-0.1cm}
\begin{align}  \label{eq:falphaw}
\nonumber
\arg \max _{\vec{\alpha, \vec{w}}} f(\vec{\alpha},\vec{w})&=\sum_{i=1}^4
\nonumber
 k_i f_i\left(\vec{\alpha},\vec{w}\right) \\
\nonumber
\text { s.t. } \quad & \alpha_i \in\{0,1\}, \\
\nonumber
& \sum_{i=1}^{\mathrm{N}} \alpha_i \leq \mathrm{M}, \\
\quad & \sum_{i=1}^M \alpha_i w_i=B,
\end{align}
\vspace{-0.1cm}
where $k_i$ is a tuning factor to weigh the importance of different objectives (and/or standardize them).

The constraints mean that the number of selected vehicles should not exceed $M$ and the utilized bandwidth by selected vehicles can not exceed the total bandwidth. To simplify this problem, we take a two-step process. We first select the vehicles solely based on their contribution to CP visual quality by solving
\vspace{-0.1cm}
\begin{align}  \label{eq:falpha}
\nonumber
\arg \max _{\vec{\alpha}} f(\vec{\alpha})&=\sum_{i=1}2 
\nonumber
 k_i f_i\left(\vec{\alpha}\right) \\
\nonumber
\text { s.t. } \quad & \alpha_i \in\{0,1\}, \\
& \sum_{i=1}^{\mathrm{N}} \alpha_i \leq \mathrm{M},
\end{align}
\vspace{-0.1cm}
and then we assign the networking resources among selected vehicles, by solving
\begin{align} \label{eq:fw}
\nonumber
\arg \max _{\boldsymbol{w}} f(\boldsymbol{w})=\sum_{i=3}^4 & k_i f_i\left(w_1, w_2, \ldots, w_M\right) \\
\nonumber
\text { s.t. } \quad & \sum_{i=1}^M p_i^{t r}=P_T, \\
\nonumber
& \sum_{i=1}^M w_i=B, \\
& E\left[D_i\right]=\frac{1}{\lambda_i} \leq T_{t r}
\end{align}
only for selected vehicles.

In characterizing bandwidth in (\ref{eq:fw}), we consider LTE-V2X, a communication technology designed for V2V communication, which supports 10 and 20 MHz channel bandwidths. This bandwidth is divided into sub-channels in the frequency domain and sub-frames in the time domain. Resource Blocks (RBs) are crucial components in the frequency domain, each consisting of 12 sub-carriers spaced by 15 kHz. The signal is formed by resource block pairs (RBPs), defined by 12 sub-carriers and carrying 14 OFDM symbols. Subchannels, defined by groups of RBPs within the same subframe, serve as the minor units of resources allocated to vehicles for transmitting Cooperative Awareness Messages (CAM). The RB, with a width of 180 kHz, becomes the smallest unit of frequency resources allocated to LTE users. Subchannels transmit data and control information, with Transport Blocks (TBs) containing complete packets, beacons, or other messages.
Resource block allocation is vital in LTE-V2X due to the need for efficient utilization of the available spectrum. The number of RBs per sub-channel can vary, and vehicles autonomously select resources without cellular infrastructure assistance.

We have used Genetic Algorithm to solve optimization problems in \ref{eq:falpha} and (\ref{eq:fw}) for its power in solving binary-vector optimization problems.

\subsection{Fusion}

  In our implementation, each vehicle utilized YOLOv8 for OD. After that, the ego vehicle captures the helpers to increase the visual range and detect objects more accurately. For the fusion part, we select the best IoU between the ego vehicle and the helper.
\vspace{-0.1cm}
\begin{align}
    IoU_s = Max(IoU_e, IoU_h)
\end{align}
\vspace{-0.1cm}
Consequently, we always select the best IoU, and the accuracy of OD will increase.

\section{Simulation}

In this section, we investigate the performance of the
proposed CP of the OD accuracy under foggy weather conditions while selecting the best vehicle under imperfect communication. In our experiments, we set the number of Vehicles
N = 10, the number of selected vehicles as M = 3, channel conditions
($\beta_i $, $\lambda_i $), relative velocity ($v_i$), vehicle position ($x_i$, $y_i$) are selected randomly.
We use Beta and exponential distributions to generate $\beta_i$ and
 $1/\lambda_i$, respectively.
We trained our model using scenarios generated by the CARLA simulator. As there is no appropriate dataset for a visual CP, we developed over 2000 images with CARLA in which there is one ego vehicle and three helpers in foggy weather. We manually annotated the dataset and then used the YOLO8 to detect pedestrians in foggy weather.

\begin{figure}[htbp]
\begin{center}
\centerline{
\includegraphics[width=0.5\columnwidth, height=0.5\columnwidth]{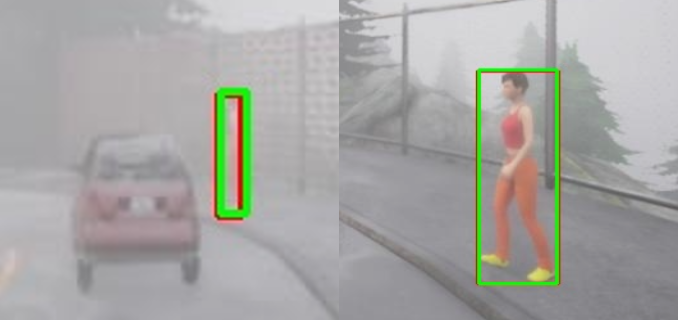}}
\caption{Objects detection by YOLOv8 to locate the pedestrian. Left: ego vehicle's view, Right: helper vehicle's view. The green box is the ground truth and the red box is the predicted bounding box by YOLOv8. }
\label{fig:pdr}
\end{center}
\vspace{-0.2 in}
\end{figure}
\begin{figure}[htbp]
\begin{center}
\centerline{
\includegraphics[width=0.8\columnwidth]{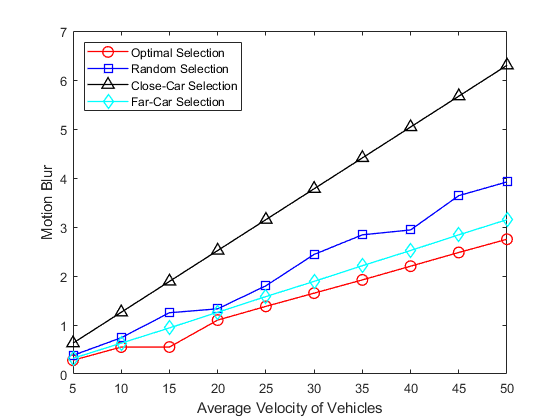}}
\caption{Optimal selection vehicle compared to other selection method. }
\label{fig:mo}
\end{center}
\end{figure}

Fig. \ref{fig:vs} shows our method performs better than other selections regarding the vehicle's visual range. We ran the selection 100 times in random selection and got the average result. Fig. \ref{fig:mo} shows that our method can select vehicles with lower motion blur, impacting image quality. 
\vspace{-0.1cm}

\begin{figure}[htbp]
\begin{center}
\centerline{
\includegraphics[width=0.8\columnwidth]{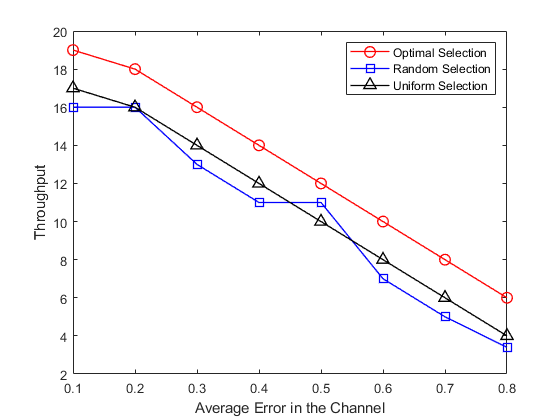}}
\caption{Optimal selection vehicle compared to random selection and Uniform selection when increasing the error versus Throughput. }
\label{fig:rb1}
\end{center}
\end{figure}

\begin{figure}[htbp]
\begin{center}
\centerline{
\includegraphics[width=0.8\columnwidth]{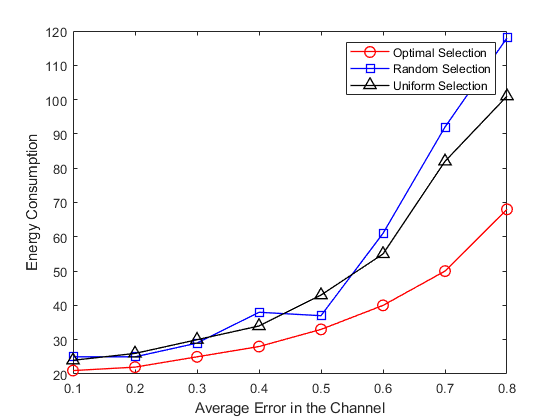}}
\caption{Optimal selection vehicle compare to random selection and Uniform selection when increase the error versus Energy Consumption. }
\label{fig:rb2}
\end{center}
\end{figure}

Fig. \ref{fig:rb1} and Fig. \ref{fig:rb2} demonstrated better performance in allocating resource blocks to vehicles in V2V communication. In random selection, resources are allocated randomly among vehicles, while in uniform allocation, resources are equally distributed among vehicles.

Table \ref{tab1} shows that when an ego vehicle recruits a helper, it can enhance OD accuracy more than when each vehicle (ego vehicle and helpers) detects objects individually. In table \ref{tab1}, we consider the scenario in a perfect communication. The result shows how much recruiting a helper can improve OD accuracy. However, we observe that Helper1 performs poorly in IoU, recall and F1 score as it 
fails to detect objects due to its distance from the ego vehicle, while the best results are achieved by recruiting Helper 2.

We introduce a packet drop rate to the images and subsequently apply the method for OD. Fig. \ref{fig:pdr} illustrates an example of the packet drop rate impact on the image. The experiment result is presented in table \ref{tab5}, with the highest accuracy for our method. The result shows how imperfect communication can impact the OD when we compare it with table \ref{tab1}.

\begin{figure}[htbp]
\begin{center}
\centerline{
\includegraphics[width=0.7\columnwidth]{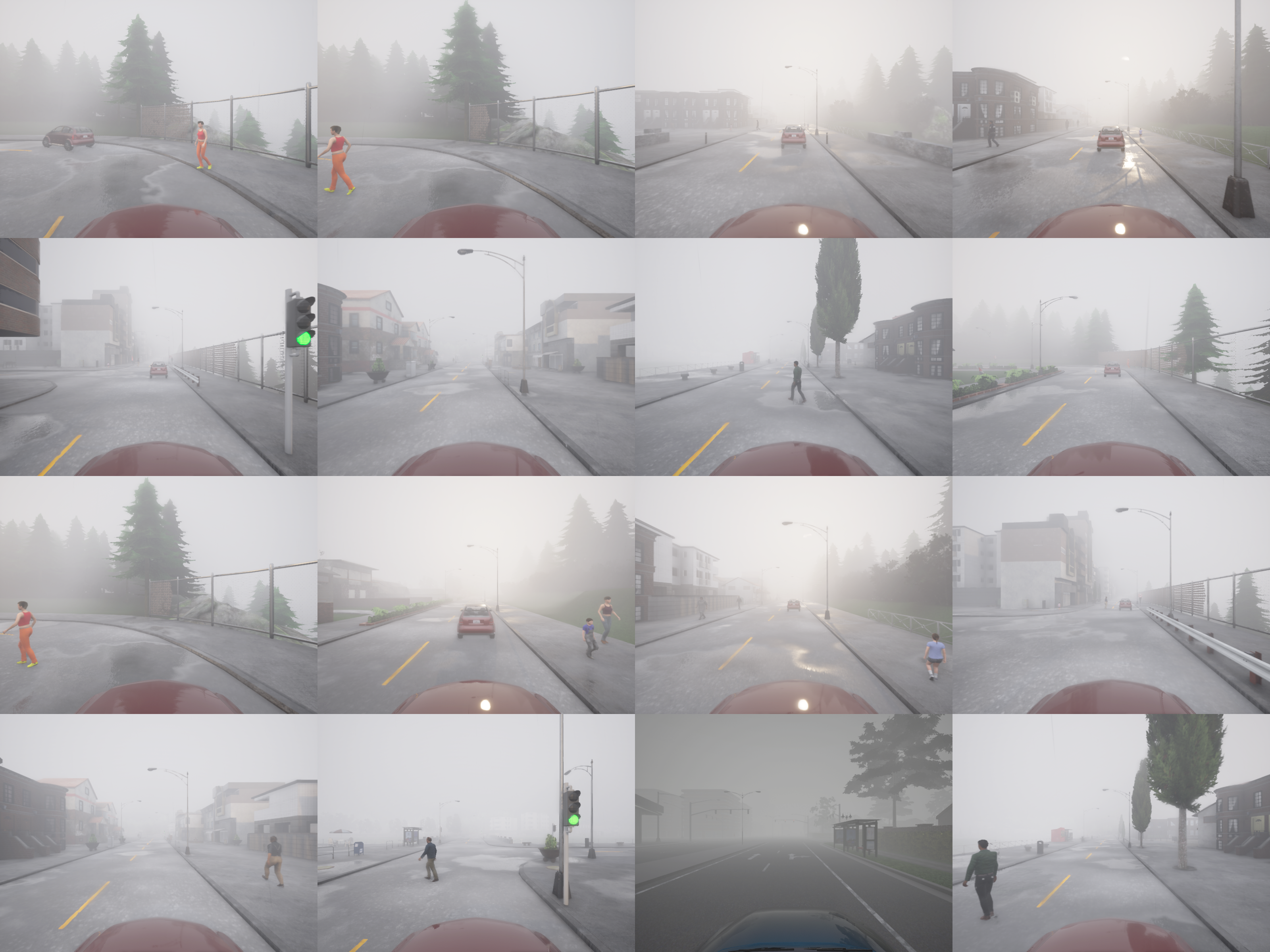}}
\caption{Data set generated with CARLA simulator}
\label{fig:vs}
\end{center}
\end{figure} 
\begin{figure}[htbp]
\begin{center}
\centerline{
\includegraphics[width=0.8\columnwidth]{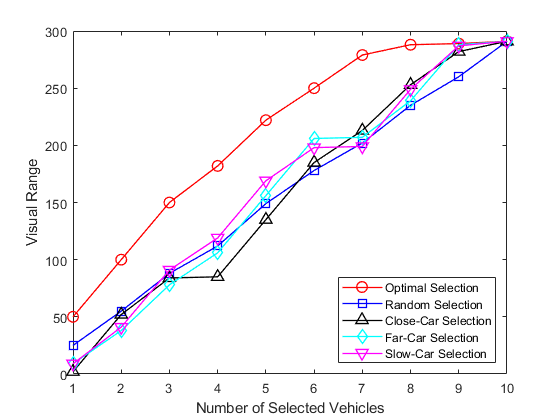}}
\caption{Optimal vehicle selection has been compared across various scenarios, including random selection (averaged over 100 runs), close-car selection (choosing the closest cars), far-car selection (choosing the farthest cars), and slow-car selection (selecting the slowest cars), all based on visual range considerations.}
\label{fig:vs}
\end{center}
\end{figure}

\vspace{-0.1cm}

\vspace{-0.1cm}

\begin{table}[htbp]
\caption{OD metrics under perfect communication.} 
\begin{center}
\begin{tabular}{ |p{0.7cm}|| p{0.45cm}|p{0.6cm}|p{0.6cm}|p{0.6cm}| p{0.6cm}| p{0.5cm}| p{0.6cm}|  }
 \hline
 & Ego & $H_{3}$ & $H_{2}$ & $H_{1}$ & Ego + $H_{3}$ & Ego + $H_{2}$  & Ego + $H_{1}$  \\
 \hline
  &  & & & & & & \\
   IoU & 0.42 & 0.68 & 0.74 & 0.43 & 0.76 & 0.86 & 0.59 \\
   
 \hline
  &  & & & & & & \\
   Recall & 0.224 & 0.343 & 0.373 & 0.208 & 0.382 & 0.435 & 0.304 \\
    \hline
  &  & & & & & & \\
   F1 Score & 0.36 & 0.501 & 0.533 & 0.347 & 0.544 & 0.591 & 0.462 \\
    \hline

 \hline
\end{tabular}
\label{tab1}
\end{center}
\end{table}

\begin{table}[htbp]
\caption{Comparison our method to other selections when we add error channel to the images} 
\begin{center}
\begin{tabular}{ |p{1.2cm}|| p{1.3cm}|p{1.3cm}|p{1.3cm}|p{1.3cm}|}
 \hline
  & Our Method & Randomly Selection & Close-Car Selection & Far-Car Selection\\
 \hline
  &  & & \\
   IoU & 0.83 & 0.55 & 0.31 & 0.49\\
   
 \hline
  &  & & \\
   Recall & 0.4 & 0.27 & 0.15 & 0.24 \\
    \hline
  &  & & \\
   F1 Score & 0.57 & 0.4 & 0.24 & 0.38 \\
    \hline

 \hline
\end{tabular}
\label{tab5}
\end{center}
\end{table}

\section{Conclusion}

Our novel approach to Cooperative Perception (CP) for Autonomous Vehicles (AVs) addresses imperfect communication. It optimizes helper selection, which leads to significant improvements in perception quality and driving safety under foggy weather. We had extensive experimentation and evaluated the validation of our method via generating an appropriate dataset in foggy weather in the CARLA simulator, demonstrating notable enhancements in CP in pedestrian detection scenarios. Our framework considered CP in imperfect communications, while most of the method works in perfect communication. Also, our method can be used for applications in AVs.


\bibliographystyle{ieeetr}
\bibliography{ref} 

\end{document}